\documentclass[runningheads]{llncs}

\usepackage{eccv}

\usepackage{eccvabbrv}

\usepackage{booktabs}
\usepackage{colortbl}
\usepackage{xcolor}
\definecolor{myyellow}{rgb}{1, 1, 0}
\definecolor{myred}{rgb}{1, 0, 0}
\definecolor{mygreen}{rgb}{0, 1, 0}
\definecolor{myblue}{rgb}{0, 0, 1}
\definecolor{mycyan}{rgb}{0, 1, 1}

\newcommand{\transp}[1]{#1!40}

\usepackage[accsupp]{axessibility}  

\usepackage[pagebackref,breaklinks,colorlinks,citecolor=eccvblue]{hyperref}

\usepackage{orcidlink}

\DeclareFontFamily{U}{stix2bb}{}
\DeclareFontShape{U}{stix2bb}{m}{n} {<-> stix2-mathbb}{}

\begin{document}

\title{Dynamic Label Injection for \\ Imbalanced Industrial Defect Segmentation} 

\titlerunning{Dynamic Label Injection for Imbalanced Industrial Defect Segmentation}

\author{Emanuele Caruso\inst{1, 2}\orcidlink{0009-0004-9693-5755}~$^*$ \and
Francesco Pelosin\inst{1}\orcidlink{0000-0002-5548-2620}~$^*$ \and 
Alessandro Simoni\inst{1}\orcidlink{0000-0003-3095-3294}~$^*$   \and
Marco Boschetti  \inst{1}
}

\authorrunning{E. Caruso, F. Pelosin, A. Simoni, M. Boschetti}

\institute{Covision Lab, Brixen, Italy \\ 
\email{\{name.surname\}@covisionlab.com} \and
Free University of Bozen-Bolzano, Bolzano, Italy \\
\email{emcaruso@unibz.it}
}

\maketitle
\def\thefootnote{*}\footnotetext{These authors contributed equally to this work.}
\begin{abstract}
  In this work, we propose a simple yet effective method to tackle the problem of imbalanced multi-class semantic segmentation in deep learning systems. One of the key properties for a good training set is the balancing among the classes. When the input distribution is heavily imbalanced in the number of instances, the learning process could be hindered or difficult to carry on. To this end, we propose a Dynamic Label Injection (DLI) algorithm to impose a uniform distribution in the input batch. Our algorithm computes the current batch defect distribution and re-balances it by transferring defects using a combination of Poisson-based seamless image cloning and \textit{cut-paste} techniques. A thorough experimental section on the Magnetic Tiles dataset shows better results of DLI compared to other balancing loss approaches also in the challenging weakly-supervised setup. The code is available at \url{https://github.com/covisionlab/dynamic-label-injection.git}.
\end{abstract}

\section{Introduction}
\label{sec:intro}
One of the most challenging problems in computer vision is to carry on the learning process of deep learning models smoothly. In particular, datasets can be apparently simple from a human perspective, but very challenging from a technical point of view of the model we are optimizing \cite{DBLP:journals/jmlr/ProbstBB19, DBLP:journals/widm/BischlBLPRCTUBBDL23}. Indeed, when setting up the training of a neural network, multiple aspects can escape human intuition and are mainly related by technicalities that are a property of the model taken into consideration \cite{vanishing}. One such case happens when the training set is heavily imbalanced, that is, the distribution of the classes is not uniform. Unfortunately, this seems to be a big challenge that dampers the learning process for a neural network, and yet it remains a simple problem for us humans. Indeed, humans can often understand a new concept from just a single example rather than needing a large number of uniformly distributed examples for each concept \cite{humans}. To this end, we aim to tackle the problem of imbalanced learning with a particular focus on industrial applications.

\begin{figure}[t]
    \centering
    \includegraphics[width=1.00\textwidth]{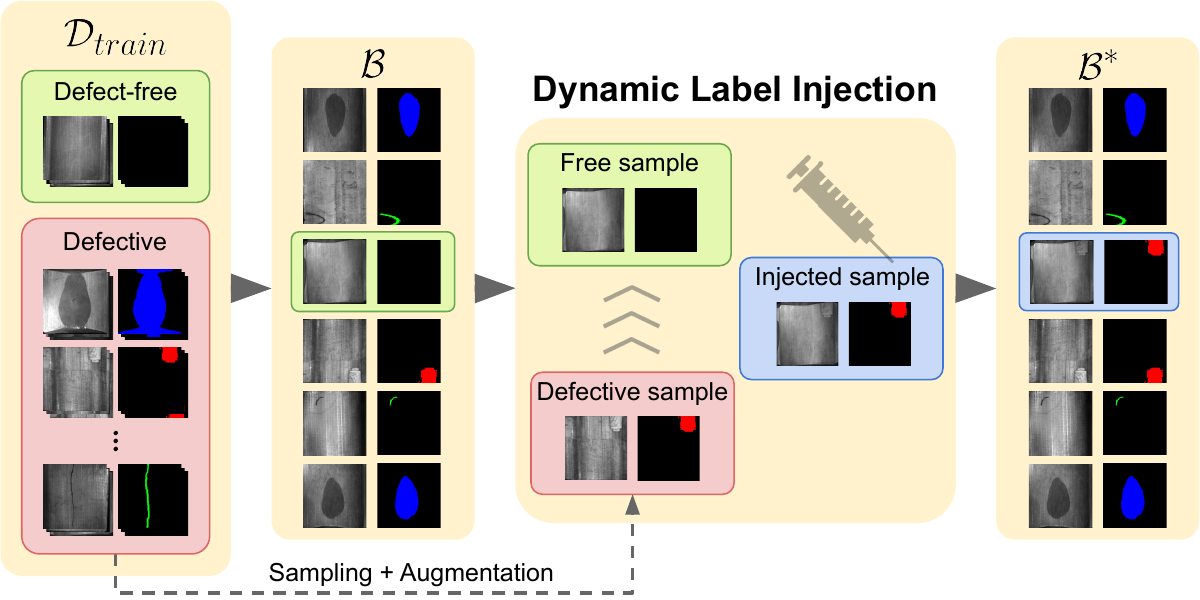}
    \caption{Overview of Dynamic Label Injection algorithm. $\mathcal{D}_{train}$ contains defect-free and defective samples. During training, defects are dynamically sampled, augmented, and injected into defect-free images to balance the batch $\mathcal{B}^{*}$, which is then fed into a multi-class defect segmentation neural network.
    }
    \label{fig:overview}
    \vspace{-1em}
\end{figure}

In the industrial setting, the typical scenario concerns a production line where the quality of the products is checked in real time. Thus, much more focus has been put by the research community into Industrial Anomaly Detection as shown in \cite{iad1, iad2, iad3, iad4} which is usually framed as single-class image classification. The great majority of benchmarks are tailored for this problem \cite{mvtec, skoloecc}. At the same time, the topic of pure segmentation and, in particular, class imbalanced segmentation in industrial settings has not been studied as deeply.


To this end, we tackle the problem of multi-class defect segmentation by proposing an algorithmic augmentation pipeline named Dynamic Label Injection (see Figure~\ref{fig:overview}), that is tailored to balance the input batch of a semantic segmentation model in an online fashion. Our approach aims to balance each input batch fed into the deep learning model by taking defect-free images and injecting augmented defects.
Standard approaches for imbalanced segmentation \cite{dice, DBLP:conf/cvpr/CuiJLSB19, focal} address the problem by proposing specific loss functions to mitigate the imbalance. Our method, instead, dynamically operates on the data, transitioning from an imbalanced to a balanced setting through an augmentation procedure, namely the Dynamic Label Injection (DLI).
We prove the effectiveness of our approach by comparing it with several alternative methods. Specifically, the pipeline has shown a significant improvement on the IoU score in the Magnetic Tiles dataset~\cite{metal_dset}. We also demonstrate the robustness of our results by performing multiple repetitions of the experiments.
Moreover, we use a combination of two techniques for the injection step: Poisson-based seamless cloning and \textit{cut-paste}. This choice has been motivated by an ablation study that evaluates the performance contribution of the two techniques. Finally, we also analyze the robustness of our online balancing algorithm in the more challenging weakly-supervised setting.

Summarizing, we propose three major contributions:
\begin{itemize}
    \item Firstly, we introduce a novel balancing algorithm, named Dynamic Label Injection, to tackle imbalanced semantic segmentation.

    \item
    Secondly, we provide an extensive comparison between our method and well-established approaches that have been proposed to tackle class imbalance.

    \item Lastly, this work represents one of the few works tackling the imbalanced multi-class defect segmentation problem with industrial data.  
    
\end{itemize}


\section{Related Works}
\label{sec:related}

\subsubsection{Industrial datasets.} The great majority of datasets used in academia drift significantly from the data that is used in the industry realm. This, in turn, results in very few standardized benchmarks for semantic segmentation. The MVTec \cite{mvtec} dataset is one of the few datasets that is consistently used in the literature and is composed of images of products and their defect mask. Each product class contains anomalous and non-anomalous examples. The MVTec dataset is mainly used for anomaly detection in industrial settings. Another frequently used dataset is the KolektorSDD2 \cite{skoloecc} which consists of different images of a single product from a visual inspection system paired with their defect mask. The challenge in this dataset comes from the variance in the defect size. Although being common benchmarks in the anomaly detection task, they are not suitable for multi-class semantic segmentation, since they contain only one defect per object.

Moreover, the Neu-Seg dataset \cite{neuseg} is an analogous dataset which, instead, is composed of more than one defect class, but our defect injection procedure requires defect-free examples and the Neu-Seg dataset does not contain any free sample. Lastly, the Magnetic Tiles dataset~\cite{metal_dset} is a common benchmark which contains images of a single product and its defect mask. This is the only dataset among the aforementioned which has defect-free examples in the multi-class segmentation setting. The dataset is heavily imbalanced and, therefore, constitutes a suitable benchmark to measure our method's performance. In the literature, this dataset has been only studied in the context of semi-supervised segmentation \cite{uap} and standard semantic segmentation in industrial settings \cite{DBLP:journals/tim/ZhangWT23a} and in CPU-based systems \cite{anet}.

\subsubsection{Semantic segmentation losses.} In semantic segmentation, loss functions play a crucial role in training models to accurately delineate object boundaries and assign correct class labels to each pixel. The Dice loss \cite{dice} maximizes the overlap between predicted and ground truth segmentation masks by calculating their intersection over union (IoU). The Lovász loss \cite{DBLP:conf/cvpr/BermanTB18}, an extension of the Jaccard index, optimizes directly for the mean IoU score, improving performance ensuring better boundary adherence. Another notable loss is the Tversky loss \cite{DBLP:conf/miccai/SalehiEG17}, a generalization of the Dice loss that adjusts for false positives and false negatives.


\subsubsection{Balancing losses.} Addressing class imbalance in machine learning is a critical challenge, and numerous strategies have been developed to mitigate its impact on model performance, mainly based on dedicated losses.

A standard technique to balance the learning process is the Weighted Cross Entropy (WCE) loss that weights the loss proportionally to the distribution of the class instances in the dataset. In the segmentation case, the distribution is calculated on the number of pixels that belong to a specific class.

Nonetheless, one of the first losses that has gained a lot of attention to solve class imbalance, is the Focal loss. Focal loss has been proposed in \cite{focal} and it was introduced to address the challenge of class imbalance in object detection tasks, where the number of background samples significantly outnumbers the number of object samples. It is a modification of the standard Cross-Entropy loss, designed to focus more on hard-to-classify examples using a parameter gamma that regulates the sensitivity. This loss can also be recast to work with pixels, unlocking the usage in imbalanced semantic segmentation problems. 


Lastly, a re-weighting scheme on the effective number of samples for each class was proposed by \cite{DBLP:conf/cvpr/CuiJLSB19}. This method introduces a Class-Balanced loss, where the loss function is re-balanced according to the effective sample size, thus mitigating the negative effects of class imbalance. This approach has been shown effective in image classification, but also in semantic segmentation tasks.



\section{Method}
\label{sec:method}
In this section, we describe the proposed online balancing algorithm used to train different multi-class defect segmentation architectures. An overview of our pipeline is depicted in Figure~\ref{fig:overview}.

\subsection{Multi-class Defect Segmentation}
In the context of multi-class defect segmentation among $C$ classes, we consider a dataset containing $N$ images along with their corresponding segmentation masks, which indicate the pixel-wise position of defects. We denote this dataset as $\mathcal{D}=\{(I_n, M_n)\}^N_{n=1}$, where $I_n \in \mathbb{R}^{3 \times H \times W}$ represents an RGB image of height $H$ and width $W$, and $M_n \in \{1, \dots, C\}^{H \times W}$ represents its segmentation mask. For each image $I_n$ in the dataset, the segmentation mask $M_n$ must satisfy the constraint that all pixels corresponding to a defect must be of the same class. Formally, this can be expressed as: $\forall n \in\{1, \ldots, N\}, \exists c \in\{1, \ldots, C\}$ such that $M_n(i, j) \in\{0, c\}$ where $M_n(i, j)$ denotes the value of the segmentation mask at pixel $(i, j)$.

An imbalanced segmentation problem arises when the dataset is not evenly distributed across the different classes. Specifically, this imbalance is characterized by significant differences in the number of examples among at least two classes. Formally, there must exist at least two classes $c$ and $c^{\prime}$ such that the number of examples $\left|\mathcal{D}_{c}\right|$ and $\left|\mathcal{D}_{c^{\prime}}\right|$ is significantly different, \ie:
\begin{equation}
|| \mathcal{D}_{c}|-| \mathcal{D}_{c^{\prime}}|| \gg 0    
\end{equation}
where $\mathcal{D}_c$ denotes the subset of the dataset containing images with defects of class $c$.

\subsection{Dynamic Label Injection} 
The process of augmenting a training batch through a \textit{cut-paste} procedure, where a label is transferred from one image to another, has been shown to be effective in self-supervised semantic segmentation as reported in \cite{DBLP:conf/cvpr/LiSYP21}. Another successful application lies in few-shot industrial segmentation as confirmed by \cite{DBLP:conf/icmcs/LinCZL21} where, in order to augment the few training examples, the procedure transfers a defect of a random image onto another one with a \textit{cut-paste} operation. 

We draw inspiration from these works to tackle imbalance defect segmentation by ensuring that each batch from dataset $\mathcal{D}$ is class-balanced during the training of our deep learning model. In particular, we use two different approaches to inject a defect in a defect-free image: Poisson-based seamless image cloning and \textit{cut-paste}. While \textit{cut-paste} simply cuts the defect in a defective image and pastes it into a defect-free image within the region defined by the mask, Poisson-based seamless cloning, described in the following, provides a more realistic blending method. Thus, we use a combination of the two by randomly applying one of them to each defect-free image in the batch. This process is repeated until a balanced batch is obtained.

\paragraph{\textbf{Poisson-based seamless image cloning.}} 
Poisson image editing~\cite{Poisson} is a technique of seamless cloning, which is the task of blending a region $\Omega$ from a source image (a defective image $I_d$) into a target image (a defect-free image $I_f$). The goal is to make the gradients of the blended region match those of $I_d$ while maintaining the boundary conditions of $I_f$, resulting in a natural transition. Formally, we solve the Poisson equation:
\label{sec:poisson-theory}
\begin{equation}
    \Delta g=\nabla \cdot \mathbf{v} \quad \text { in } \Omega
\end{equation}
where $\Delta$ is the Laplace operator and $\mathbf{v}=\nabla I_d$ is the gradient of the source image. The boundary conditions are:
\begin{equation}  
g=I_f \quad \text { on } \quad \partial \Omega
\end{equation}
where $\partial \Omega$ is the boundary of $\Omega$. The solution $g$ minimizes the energy functional:
\begin{equation}  
E(g)=\int_{\Omega}\|\nabla g-\nabla I_f\|^2 dx
\end{equation}
This approach ensures that the blended region $g$ has a smooth gradient transition and integrates seamlessly into the target image $I_f$. Poisson image editing is particularly effective for object insertion, background replacement, and texture flattening.

\paragraph{\textbf{Dynamic Label Injection.}} 
Given a sufficiently large batch $\mathcal{B}=\left\{I_1, \ldots, I_{|\mathcal{B}|}\right\}$ that contains both defect-free images $I_f$ and defective images $I_d$, we aim to augment the defect-free images to compensate for the class imbalance obtaining a new batch $\mathcal{B}^*$. This involves properly transferring defects present in the dataset onto the defect-free images through our Dynamic Label Injection algorithm.

The balancing process consists of iterating over the defect-free images in the batch and determining the class of the defect to inject based on the current imbalance. Specifically, we inject a defect from the class that has the fewest occurrences in the batch at the current iteration. We stop the balancing procedure once the class distribution within the batch is uniform.

\paragraph{\textbf{Algorithmic formulation.}} Let $\mathcal{B}_d$ and $\mathcal{B}_f$ denote the sets of defective and defect-free images in $\mathcal{B}$, respectively. For each $I_f \in \mathcal{B}_f$:
\begin{enumerate}
    \item[a)] Determine the class $c^*$ with the fewest occurrences in the current batch $\mathcal{B}$:

    \begin{equation}
    c^*=\arg \min _{c \in\{1, \ldots, C\}}\left|\left\{I_d \in \mathcal{B}_d \vert M_d \in \{ 0, c \}\right\}\right|    
    \end{equation}
    
    where $M_d$ is the segmentation mask of the defect's class $c$.
    
    \item[b)] Randomly sample a defective image $I^{\prime}_d \in \mathcal{D}_{\text {train }}$ with its segmentation mask $M^{\prime}_d \in \{ 0, c^* \}$.

    \item[c)] Transform $I^{\prime}_d$ applying random rotation, flip and translation operations.

    \item[d)] Randomly select a function between \textit{cut-paste} and Poisson-based seamless image cloning, to be used as an injection function.
    
    \item[e)] Inject the defect in $I^{\prime}_d$ into the defect-free image $I_f$, leading to $I_f^*$. 
\end{enumerate}
A pseudo-code of the algorithm is presented in Algorithm~\ref{alg:balancing}.

\begin{algorithm}[t]
\caption{Dynamic Label Injection algorithm. Comments in \textcolor{blue}{blue}}
\label{alg:balancing}
\begin{algorithmic}[1]
\REQUIRE Batch $\mathcal{B}$, Dataset $\mathcal{D}$
\ENSURE Balanced Batch $\mathcal{B}^*$

\STATE Initialize $\mathcal{B}^* \gets \mathcal{B}$
\STATE Identify defect-free images $\mathcal{B}_f$ from $\mathcal{B}$
\STATE Identify defective images $\mathcal{B}_d$ from $\mathcal{B}$

\WHILE{$\mathcal{B}$ is imbalanced}
    \FOR{each $I_f \in \mathcal{B}_f$}
        
        \STATE $c^* \gets \arg \min_{c \in \{1, \dots, C\}} \left| \{I_d \in \mathcal{B}_d \vert M_d \in \{ 0, c \} \} \right|$ \hfill \textcolor{blue}{\COMMENT{Step (a)}}
        
        \STATE $I^\prime_d : \{ \mathcal{D}_{\textit{train}} \vert M^\prime_d \in \{ 0, c^* \} \}$ \hfill \textcolor{blue}{\COMMENT{Step (b)}}
        
        \STATE $I^\prime_d \gets \textbf{Transform}(I^\prime_d)$ \hfill \textcolor{blue}{\COMMENT{Step (c)}}
        
        \STATE $\textbf{InjectionFunction} : \{\textit{Poisson}, \textit{cut-paste}\}$ \hfill \textcolor{blue}{\COMMENT{Step (d)}}

        \STATE $I_f^* \gets \textbf{InjectionFunction}(I_f, I^\prime_d)$ \hfill \textcolor{blue}{\COMMENT{Step (e)}}
        
        \STATE Update $\mathcal{B}^*$ with $I_f^*$ 
    \ENDFOR
\ENDWHILE
\RETURN $\mathcal{B}^*$
\end{algorithmic}
\end{algorithm}

\subsection{Losses}
We use Dice and Binary Cross Entropy (BCE) losses to handle imbalanced multi-class semantic segmentation achieving accurate predictions. Given an image $I_n$, the ground truth $M_n$, and a predicted mask $\hat{M}_n$, we employ a Binary Cross Entropy loss to assess the pixel-wise prediction error:
\begin{equation}
    \mathrm{BCE}=-\frac{1}{H \times W} \sum\left(M_n \log \hat{M}_n+\left(1-M_n\right) \log \left(1-\hat{M}_n\right)\right)
\end{equation}

We also use Dice loss, which measures the overlap between the predicted mask $\hat{M}_n$ and the ground truth $M_n$:
\begin{equation}
\text { Dice Loss }=1-\frac{2 \sum\left(\hat{M}_n M_n\right)}{\sum \hat{M}_n+\sum M_n}
\end{equation}
Dice loss focuses on maximizing overlap, making it robust against class imbalance, while BCE ensures pixel-wise accuracy. Combining these losses improves the segmentation model performance.
\section{Experiments}
\label{sec:experiments}
In this section, we validate the proposed method on a highly imbalanced industrial dataset with images of magnetic tile defects. After introducing the dataset in Section~\ref{sec:dataset}, we explain our experimental setting in Section~\ref{sec:exp-setting}. We compare our results with other class balancing losses in Section~\ref{sec:results}, providing quantitative and qualitative evaluations. We also examine performance robustness in the challenging weakly-supervised setting with less training data in Section~\ref{sec:robustness}. Moreover, an ablation study is conducted in Section~\ref{sec:ablation} to compare the combined injection method with the individual use of Poisson or \textit{cut-paste} strategy. 

\begin{figure}[t]
    \centering
    \begin{minipage}{0.48\textwidth}
        \centering
        \setlength{\tabcolsep}{3pt}
        \resizebox{0.95\textwidth}{!}{
        \begin{tabular}{ccc}
            \toprule
            \multicolumn{3}{c}{\textbf{Magnetic Tiles Dataset}} \\
            \midrule
            & \textit{N. Images} & \textit{Defect Area \%} \\
            \textbf{Defect-free} & 952 & - \\
            
            \rowcolor{\transp{myyellow}}
            \textbf{Blowhole} & 115 & 00.21 \scriptsize{$\pm$ 00.16}   \\
            
            \rowcolor{\transp{myred}}
            \textbf{Break} & 85  & 04.24 \scriptsize{$\pm$ 07.22}   \\

            \rowcolor{\transp{mygreen}}
            \textbf{Crack} & 57  & 00.49 \scriptsize{$\pm$ 00.53}   \\

            \rowcolor{\transp{myblue}}
            \textbf{Fray} & 32  & 17.10 \scriptsize{$\pm$ 14.15} \\

            \rowcolor{\transp{mycyan}}
            \textbf{Uneven} & 103 & 23.90 \scriptsize{$\pm$ 15.32} \\ 
            \bottomrule
        \end{tabular}
        }
        \label{tab:dataset}
    \end{minipage}
    \begin{minipage}{0.48\textwidth}
        \centering
        \includegraphics[width=\textwidth]{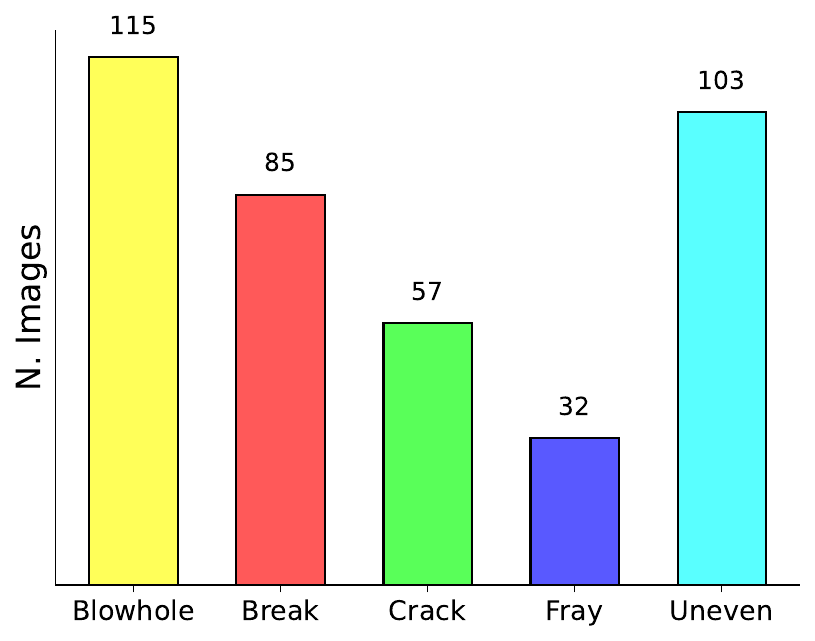}
        \label{fig:hist}
    \end{minipage}
    \vspace{-2em}
    \caption{Magnetic Tiles dataset class imbalanced distribution.}
    \vspace{-1.3em}
    \label{fig:dataset-distr}
\end{figure}

\subsection{Dataset}
\label{sec:dataset}
The Magnetic Tiles (MT) dataset~\cite{metal_dset} contains 1344 images where 952 are labeled as free and 392 as defective. There are 5 classes of defects (blowhole, break, crack, fray, uneven) which differ a lot in terms of shape and mask pixel occupation with respect to the full image resolution. Moreover, this dataset is well-suited for studying the class imbalance problem in the industrial semantic segmentation domain. As can be seen in Figure~\ref{fig:dataset-distr}, the defect classes are highly imbalanced, so our method can leverage the free images as additional data augmented by injecting defects. The set of transformations applied to each defect before the injection consists of a sequence of random flip, scaling, shifting, and rotating operations. 

\begin{table}[t]
\centering
\caption{Quantitative comparison of all balancing methods for different lightweight encoders of the UNet. We report mean IoU and standard deviation on 5 different seeds.}
\label{tab:results}
\setlength{\tabcolsep}{3pt} 
\resizebox{\textwidth}{!}{
\begin{tabular}{r|ccc|c}
    \toprule
    \textbf{Method} & $\ell$\textbf{-ResNest50d \scriptsize{(1.6M)}} &  $\ell$\textbf{-ResNet18 \scriptsize{(0.7M)}} &  $\ell$\textbf{-MobileoneS1 \scriptsize{(0.3M)}} & \textbf{Avg} \\
    \midrule
    Baseline & 0.774 \scriptsize{$\pm$ 0.039} & 0.746 \scriptsize{$\pm$ 0.018}  & 0.743 \scriptsize{$\pm$ 0.021} & 0.754 \scriptsize{$\pm$ 0.014} \\
    Focal loss \cite{focal}  & 0.726 \scriptsize{$\pm$ 0.066} & 0.658 \scriptsize{$\pm$ 0.067} & 0.683 \scriptsize{$\pm$ 0.015} & 0.689 \scriptsize{$\pm$ 0.028} \\
    Balanced loss \cite{DBLP:conf/cvpr/CuiJLSB19}  & 0.778 \scriptsize{$\pm$ 0.018} & 0.723 \scriptsize{$\pm$ 0.018} & 0.733 \scriptsize{$\pm$ 0.024} & 0.745 \scriptsize{$\pm$ 0.024} \\
    WCE loss  & 0.784 \scriptsize{$\pm$ 0.016} & 0.741 \scriptsize{$\pm$ 0.016} & 0.740 \scriptsize{$\pm$ 0.019} & 0.755 \scriptsize{$\pm$ 0.021} \\
    \midrule
    \textbf{DLI}  & \textbf{0.831 \scriptsize{$\pm$ 0.018}} & \textbf{0.780 \scriptsize{$\pm$ 0.016}} & \textbf{0.789 \scriptsize{$\pm$ 0.016}} & \textbf{0.800 \scriptsize{$\pm$ 0.022}} \\
    \bottomrule
\end{tabular}
}
\end{table}

\begin{figure}[t]
    \centering
    \includegraphics[width=\linewidth]{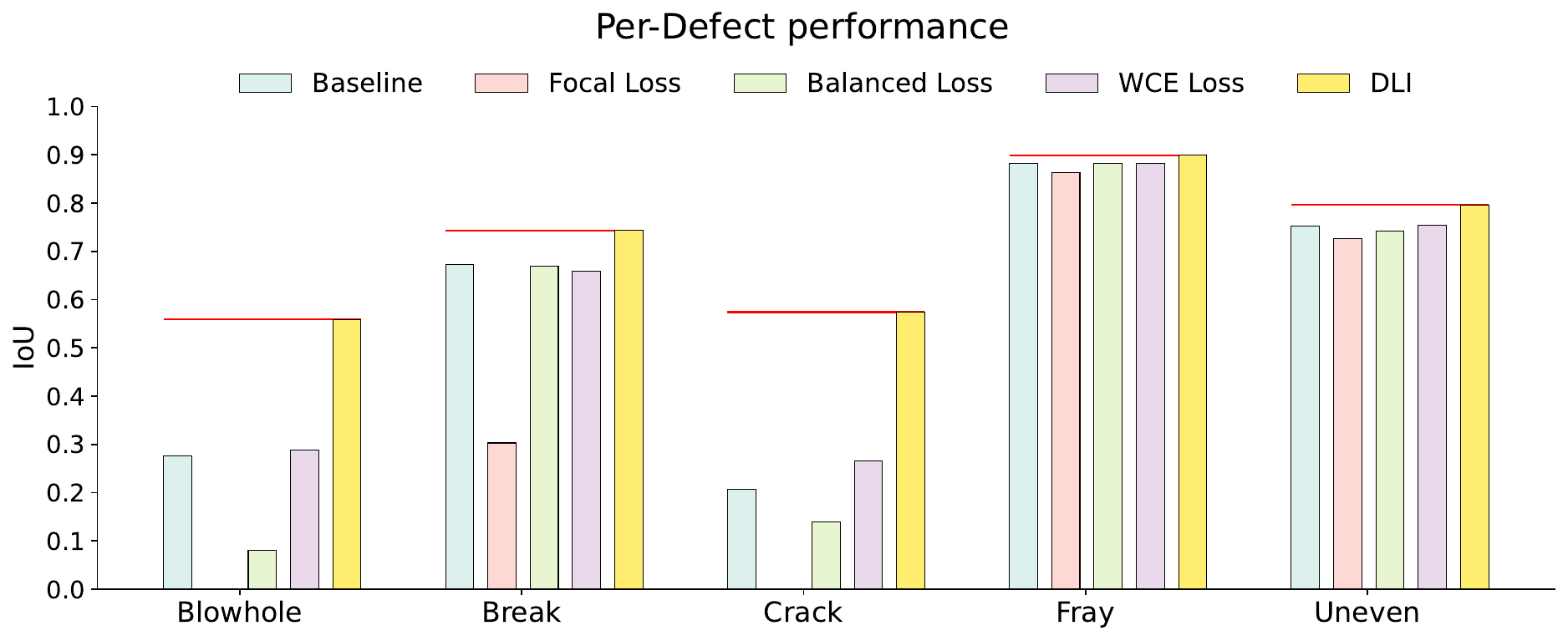}
    \caption{IoU scores for each defect averaged over 5 different seeds. An horizontal red line highlights the gap between DLI and competitors.}
    \vspace{-1em}
    \label{fig:results-perclass}
\end{figure}

\begin{figure}[t]
    \centering
    \includegraphics[width=0.9\linewidth]{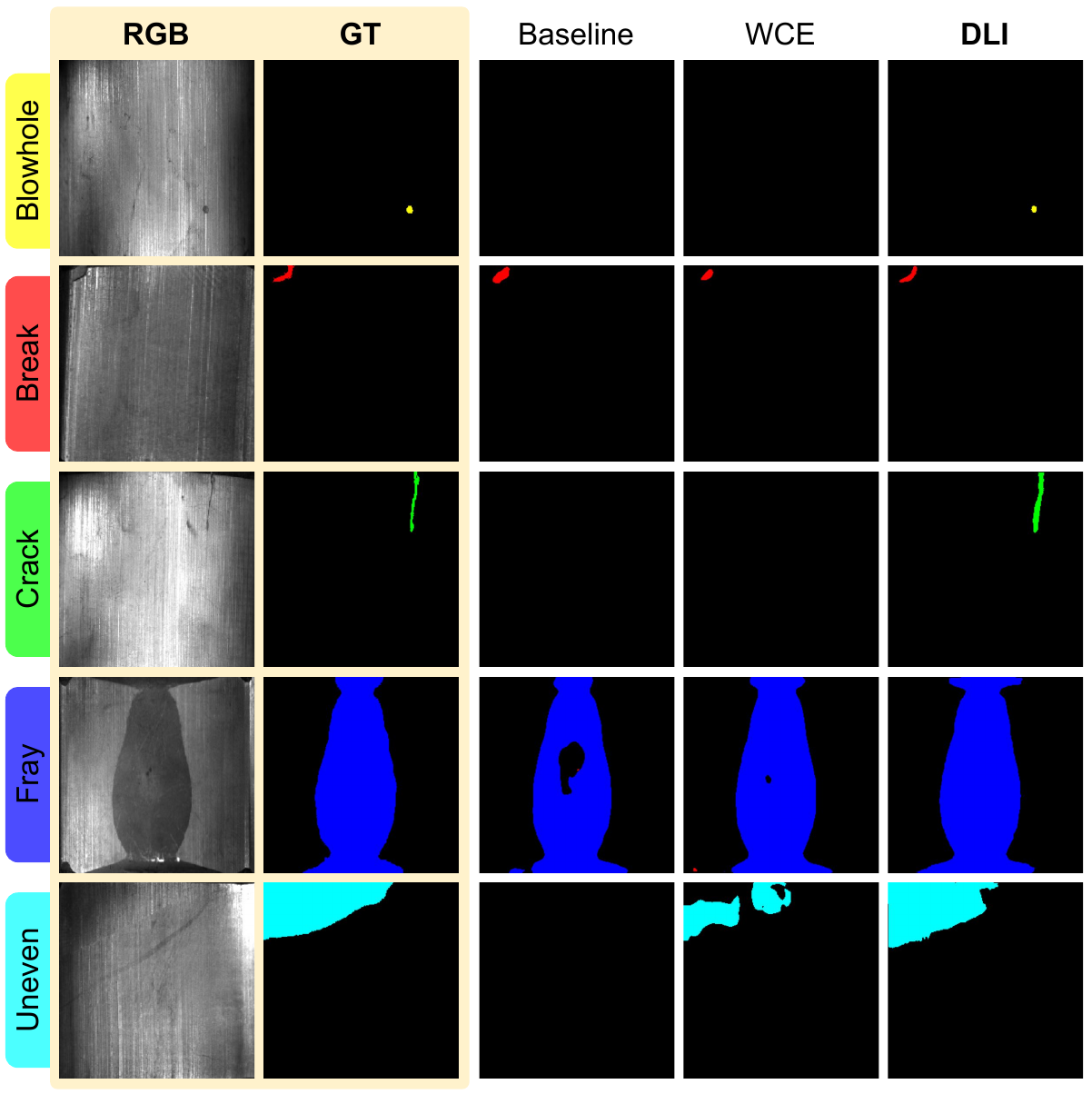}
    \caption{Qualitative samples for each class of defects in the Magnetic Tiles dataset.}
    \label{fig:qualitative-results}
    \vspace{-1em}
\end{figure}

\subsection{Experimental setting}
\label{sec:exp-setting}
The network used for the semantic defect segmentation tasks is a UNet~\cite{unet} architecture. In industry, memory occupation is a highly important requirement. To this end, we devised different lightweight backbones used as encoders. We selected ResNest50d~\cite{resnest}, ResNet18~\cite{resnet}, and MobileoneS1~\cite{mobilone} removing the last two stages of each network, significantly decreasing the number of trainable parameters. We use the prefix $\ell$- as lightweight to stress this alteration. Each model takes as input an RGB image resized to a dimension of $256 \times 256$ and outputs a set of segmentation masks $C \times 256 \times 256$ with $C = 5$, which is the number of defect classes in the dataset. All models start from pretrained ImageNet weights and are trained for $1000$ epochs using the Adam~\cite{adam} optimizer, a cosine annealing scheduler spanning over the number of epochs and a learning rate of $5e^{-4}$. We split the dataset into 80\% for the training set and 20\% for the validation set. We repeat the experiments over 5 different seeds as cross-validation.

We compare DLI against several class balancing losses: Binary Cross Entropy (referred as Baseline), Focal \cite{focal}, Balanced \cite{DBLP:conf/cvpr/CuiJLSB19} and Weighted Cross Entropy (WCE). All the approaches have been coupled with a Dice loss without any label injection and see also the defect-free images during training for a fair comparison.

\subsection{Results}
\label{sec:results}

\begin{figure}[t]
    \centering
    \includegraphics[width=\textwidth]{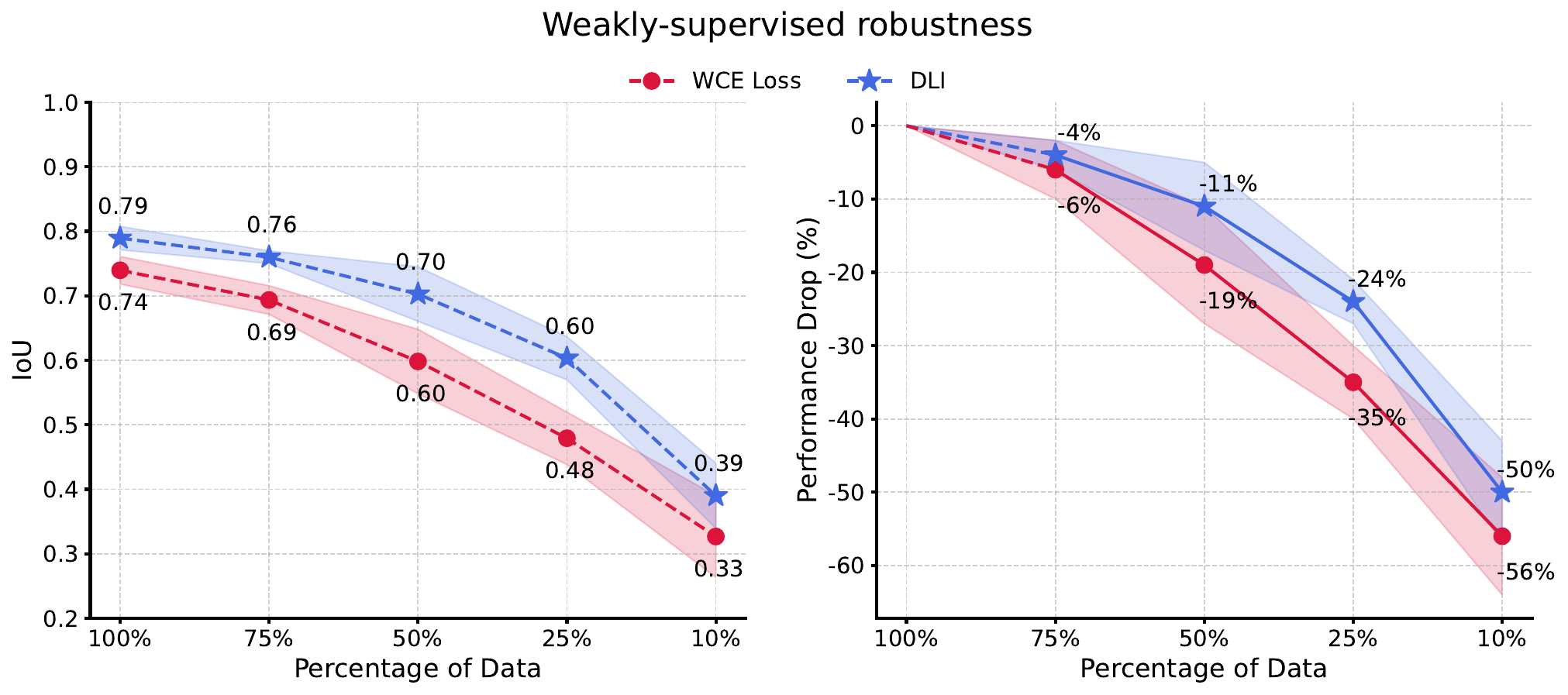}
    \caption{The plot shows how IoU varies with the percentage of data in the training set, using DLI and WCE loss. The left chart displays absolute IoU values, while the right chart shows the performance drop compared to training with the full dataset.}
    \label{fig:weakly}
    \vspace{-1em}
\end{figure}

\begin{figure}[!ht]
    \centering
    \includegraphics[width=0.98\linewidth]{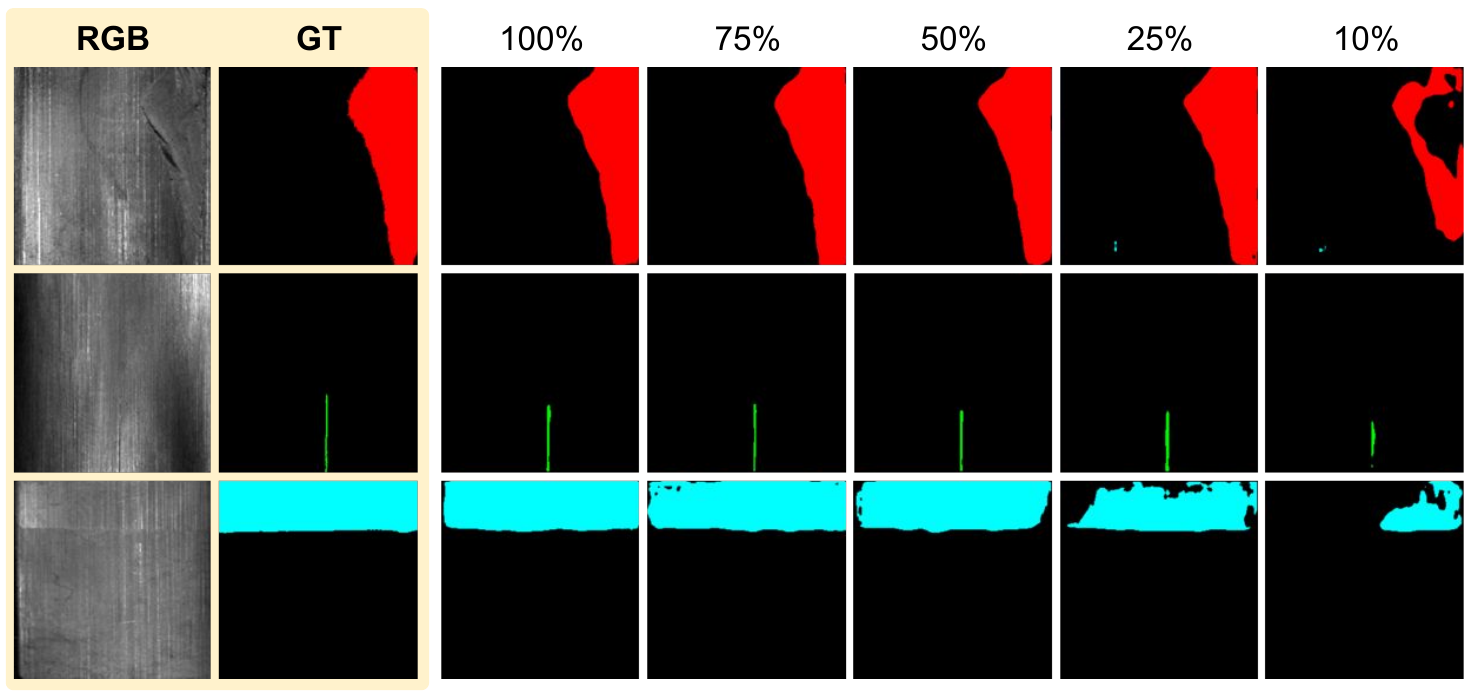}
    \caption{Qualitative results of DLI in the challenging weakly-supervised setting.}
    \label{fig:qualitative-weakly}
    \vspace{-1em}
\end{figure}

\subsubsection{Quantitative results.} Table~\ref{tab:results} presents statistics on the Intersection over Union (IoU) value obtained during training, computed on the validation set for each considered model. Formally, we computed the IoU with the following equation:
\begin{equation}
    \label{eq:iou}
    \text{IoU}_{\mathcal{D}_{val}}= \frac{\text{TP}_{\mathcal{D}_{val}}}{\text{TP}_{\mathcal{D}_{val}}+\text{FN}_{\mathcal{D}_{val}}+\text{FP}_{\mathcal{D}_{val}}}
\end{equation}
where $\text{TP}_{\mathcal{D}_{val}},\text{FN}_{\mathcal{D}_{val}},\text{FP}_{\mathcal{D}_{val}}$ are all the true positives, false negatives and false positives predicted on the validation set $\mathcal{D}_{val}$.
We report the mean and standard deviation computed with the 5 different seeds, each of which also influences the train-validation data split. As can be observed from Table~\ref{tab:results}, our DLI method outperforms all competitors for each different network.

Moreover, in Figure~\ref{fig:results-perclass}, we also provide the IoU values for each class and method, grouping with different defect types. It can be observed that the DLI method provides better IoU values for every defect class, resulting also in less class imbalance in terms of performance. In particular, the performance gap is notable on defects with a small occupation area, \eg blowhole and crack.

\begin{table}[t]
\centering
\caption{Ablation study that compares \textit{cut-paste} only, Poisson-based seamless cloning only, and the combination between the two. Average and standard deviation on 5 different seeds.}
\label{tab:lfp-vs-lfs}
\setlength{\tabcolsep}{3pt}
\resizebox{\textwidth}{!}{
    \begin{tabular}{cc|ccc|c}
        \toprule
        \textbf{Cut-Paste} & \textbf{Poisson} & $\ell$\textbf{-ResNest50d \scriptsize{(1.6M)}} &  $\ell$\textbf{-ResNet18 \scriptsize{(0.7M)}} &  $\ell$\textbf{-MobileoneS1 \scriptsize{(0.3M)}} & \textbf{Avg} \\ \midrule
        \checkmark &  & 0.818 \scriptsize{$\pm$ 0.015} & 0.776 \scriptsize{$\pm$ 0.013} & 0.779 \scriptsize{$\pm$ 0.011} & 0.791 \scriptsize{$\pm$ 0.019} \\
         & \checkmark & 0.826 \scriptsize{$\pm$ 0.011} & 0.769 \scriptsize{$\pm$ 0.015} & 0.769 \scriptsize{$\pm$ 0.010} & 0.788 \scriptsize{$\pm$ 0.027}\\
        \checkmark & \checkmark & \textbf{0.831 \scriptsize{$\pm$ 0.018}} & \textbf{0.780 \scriptsize{$\pm$ 0.016}} & \textbf{0.789 \scriptsize{$\pm$ 0.016}} & \textbf{0.800 \scriptsize{$\pm$ 0.022}} \\ \bottomrule
    \end{tabular}
}
\end{table}

\begin{figure}[t]
    \centering
    \includegraphics[width=0.95\linewidth]{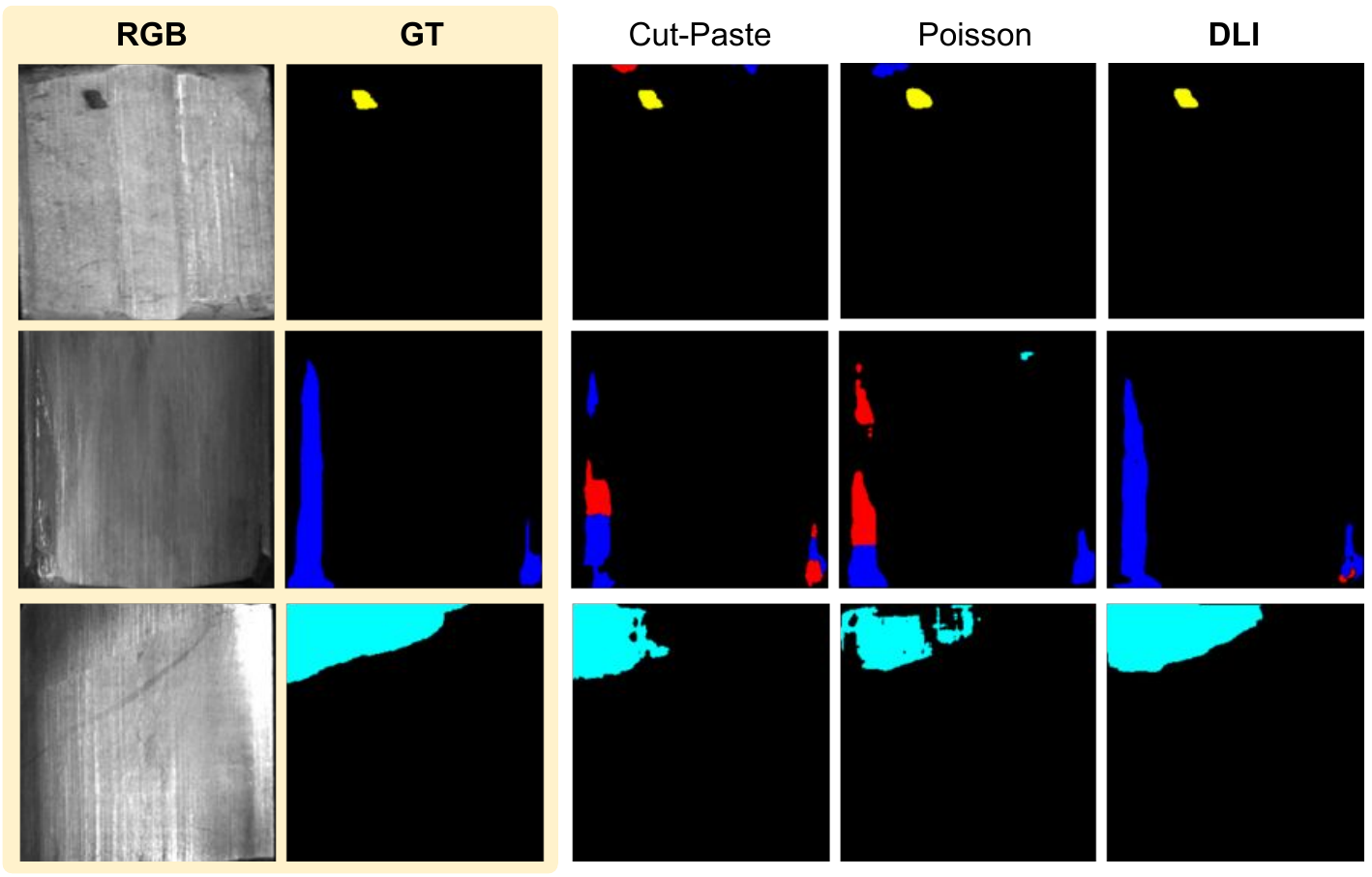}
    \caption{Qualitative results of the ablation study using \textit{cut-paste}, \textit{Poisson} or DLI.}
    \label{fig:qualitative-ablation}
    \vspace{-1em}
\end{figure}

\subsubsection{Qualitative results.} In Figure~\ref{fig:qualitative-results}, we show a qualitative comparison of all the classes of the Magnetic Tiles dataset. We choose the baseline approach and the best balancing loss technique to compare our DLI method. As can be seen, DLI provides a higher accuracy on the defects' segmentation task, while the other methods generate holes in the predicted masks or even fail to detect smaller defects.

\begin{figure}[t!]
    \centering
        \includegraphics[width=0.95\textwidth]{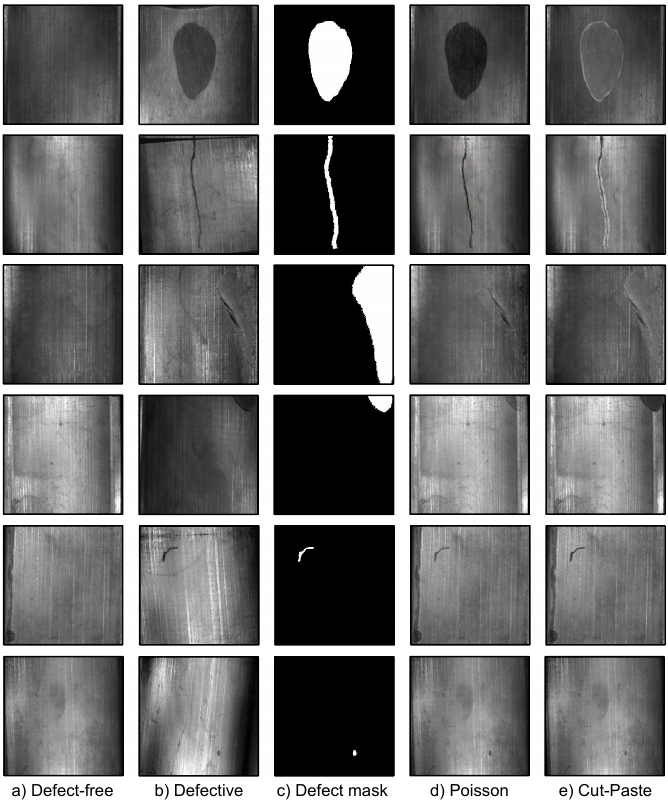}
    \caption{Qualitative comparison between Poisson and \textit{cut-paste} label injection. Each column shows defect-free images (a) to be augmented, sampled defective images (b), defect masks (c), defects injected using Poisson (d) and \textit{cut-paste} (e).}
    \label{fig:lfp-vs-lfs}
    \vspace{-1em}
\end{figure}

\subsection{Weakly-supervised robustness}
\label{sec:robustness}
We further conducted a study to demonstrate the higher robustness of our method when dealing with less data, compared to other methods. For this purpose, we compared our DLI method with the one based on the Weighted Cross-Entropy (WCE) loss, training a model sampling for each class the 10\%, 25\%, 50\%, 75\%, and 100\% of the available training data. We choose the WCE method because it is the best competitor in Table~\ref{tab:results}. The experiments were conducted using the $\ell$-MobileoneS1~\cite{mobilone} encoder as the backbone architecture since it provides a lightweight network, well-suited for industrial applications.
By looking at Figure~\ref{fig:weakly}, it can be observed how the performance decreases in the two methods when using a reduced amount of data during training. This outcome denotes that our approach has a smoother drop compared to the WCE loss, showing a smaller performance loss even with significant data reductions. This is due to the great variability that our proposal introduces when injecting each batch. Each injection creates never-seen-before examples able to unlock overall better generalization capabilities. As can be seen, there is just a $-4\%$ performance drop with $75\%$ of the data available and a $-11\%$ performance drop when the data is halved. These results make our proposal a good candidate to tackle weakly-supervised problems. Some qualitative samples of our results are depicted in Figure~\ref{fig:qualitative-weakly}.

\subsection{Ablation study}
\label{sec:ablation}
In this ablation study, we demonstrate the impact of including the Poisson-based seamless cloning method compared to the \textit{cut-paste} method. Specifically, we show the performance of our DLI algorithm using only Poisson-based seamless cloning, only \textit{cut-paste}, and the combination between the two, by randomly choosing one or the other for each batch injection. We provide a quantitative and qualitative comparison in Table~\ref{tab:lfp-vs-lfs} and Figure~\ref{fig:qualitative-ablation} respectively. As can be seen, the combined approach, which represents our final method, has the best performance among the different networks. The intuition is that for some defects it is more important to have a smooth transition between the background and the defect, while in some other cases, it is more important to have a strong perceptual difference. To achieve this, random alternation between the two methods provides the best results. In Figure~\ref{fig:lfp-vs-lfs}, we also offer a qualitative comparison between Poisson and \textit{cut-paste} label injection. Specifically, the first two rows highlight the advantage of using Poisson-based seamless cloning, which provides more consistency in terms of average intensity values compared to the \textit{cut-paste} method. The two rows in the middle provide an example where the pixel replacement of \textit{cut-paste} is preferred to the Poisson edge blending, as the defect represents a physical replacement in a big portion of the image. On the other hand, the last two rows provide a case where both \textit{cut-paste} and Poisson provide a good solution.
\section{Conclusions}
\label{sec:conclusions}
This paper introduces Dynamic Label Injection (DLI), a novel method to address imbalanced multi-class semantic segmentation problems. By dynamically balancing input batches using Poisson-based seamless image cloning and \textit{cut-paste} techniques, our approach ensures a uniform class distribution, enhancing the learning process and improving model performance. Experiments on the Magnetic Tiles dataset demonstrate the effectiveness of our method, showing superior IoU scores compared to other class balancing techniques.
We further confirm the robustness of DLI, maintaining high performance in the challenging weakly-supervised setting and when individually using Poisson or \textit{cut-paste} to inject the defects. Thus, the contribution of including the Poisson-based seamless image cloning technique together with \textit{cut-paste} has been demonstrated in its respective ablation study. Our approach consistently outperforms competitors' approaches over various model architectures verified by several experimental runs with different seeds.

%
%
\bibliographystyle{splncs04}
\bibliography{main}

\end{document}